\documentclass[10pt,twocolumn,letterpaper]{article}

\usepackage{iccv}
\usepackage{times}
\usepackage{epsfig}
\usepackage{graphicx}
\usepackage{amsmath}
\usepackage{amssymb}
\usepackage{booktabs}

\usepackage{algorithm}
\usepackage{algorithmicx}
\usepackage{algpseudocode}
\usepackage{diagbox}
\usepackage{multirow}
\usepackage{bm}
\usepackage{color}
\usepackage[table,xcdraw,dvipsnames]{xcolor}
\definecolor{darkspringgreen}{rgb}{0.0, 0.45, 0.05}
\definecolor{carmine}{rgb}{0.68, 0.05, 0.0}
\definecolor{iODBlue}{RGB}{220, 232, 250}
\usepackage{pifont}
\usepackage{mathtools}

\usepackage[accsupp]{axessibility} 

\usepackage[breaklinks=true,bookmarks=false]{hyperref}

\usepackage[capitalize]{cleveref}
\crefname{section}{Sec.}{Secs.}
\Crefname{section}{Section}{Sections}
\Crefname{table}{Table}{Tables}
\crefname{table}{Tab.}{Tabs.}

\iccvfinalcopy 

\def\iccvPaperID{9684} 

\ificcvfinal\fi

\newcommand{\cmark}{\ding{51}\xspace}%
\newcommand{\xmark}{\ding{55}\xspace}%

\begin{document}

\title{DiffDis: Empowering Generative Diffusion Model with Cross-Modal Discrimination Capability}

\author{
Runhui Huang$^{1}$~
Jianhua Han$^2$~ 
Guansong Lu$^2$~
Xiaodan Liang$^1 \thanks{Corresponding author: xdliang328@gmail.com}$~\\
Yihan Zeng$^2$~
Wei Zhang$^2$~
Hang Xu$^2$ \\
$^1$Shenzhen Campus of Sun Yat-sen University~
$^2$Huawei Noah's Ark Lab\\
}

\maketitle
\ificcvfinal\thispagestyle{empty}\fi

\begin{abstract}
{
Recently, large-scale diffusion models, e.g., Stable diffusion and DallE2, have shown remarkable results on image synthesis. 
On the other hand, large-scale cross-modal pre-trained models (e.g., CLIP, ALIGN, and FILIP) are competent for various downstream tasks by learning to align vision and language embeddings. In this paper, we explore the possibility of jointly modeling generation and discrimination. Specifically, we propose \textbf{DiffDis} to unify the cross-modal generative and discriminative pretraining into one single framework under the diffusion process. DiffDis first formulates the image-text discriminative problem as a generative diffusion process of the text embedding from the text encoder conditioned on the image. Then, we propose a novel dual-stream network architecture, which fuses the noisy text embedding with the knowledge of latent images from different scales for image-text discriminative learning. 
Moreover, the generative and discriminative tasks can efficiently share the image-branch network structure in the multi-modality model. 
Benefiting from diffusion-based unified training, DiffDis achieves both better generation ability and cross-modal semantic alignment in one architecture. Experimental results show that DiffDis outperforms single-task models on both the image generation and the image-text discriminative tasks, e.g., 1.65\% improvement on average accuracy of zero-shot classification over 12 datasets and 2.42 improvement on FID of zero-shot image synthesis.
}
\end{abstract}

\section{Introduction} \label{sec:intro}

\begin{figure}[t!]
\begin{center}
\includegraphics[width=0.95\linewidth]{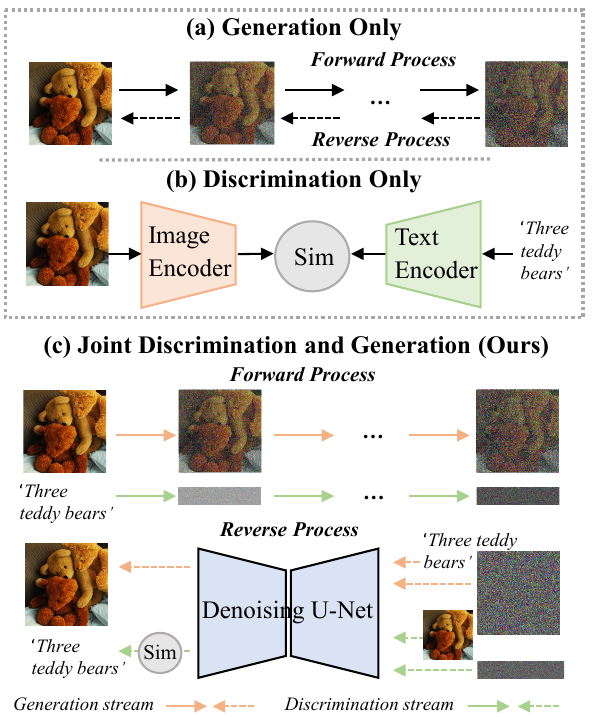}
\end{center}
\caption{Comparison of our framework and single-task models. (a) The diffusion-based image generation-only model. (b) The image-text discrimination-only model. (c) Our DiffDis joints the discriminative and generative tasks under the diffusion processing into one framework. Better viewed in colors.
}
\label{fig:introduction}
\end{figure}

\textit{“What I cannot create, I do not understand.”} by Richard Feynman (a well-known theoretical physicist).

Recently, large-scale diffusion models (DM)~\cite{ddpm,diffusion,yang2022diffusion}  such as Stable Diffusion~\cite{rombach2022high} and DallE2~\cite{ramesh2022hierarchical}  have shown impressive
results in image synthesis and re-define the capacity of state-of-the-art text-guided image synthesis.
Typically speaking, these models contain more than one billion parameters and have a large model capacity; thus have a good generalization and cover an extensive range of domains.
Here, by rethinking the famous remarks from Richard Feynman, we explore \textit{whether such powerful generative models can learn the ability to further discriminate and understand cross-modal data}.

On the other hand, recent large-scale Vision-Language Pre-training (VLP) models~\cite{radford2021learning, jia2021scaling, yao2022filip, li2020unimo} like~CLIP\cite{radford2021learning} and ALIGN~\cite{jia2021scaling} have 
demonstrated success in various downstream zero-shot image classification or retrieval tasks. 
Similar to large-scale diffusion models, these models are pre-trained with millions of image-text pairs collected from the Internet. 
The critical idea of these works is to contrastively align the image and text embeddings into a joint feature space, thus gaining zero-shot discrimination capability, which is different from the diffusion models that consider the problem as the parameterized Markov chain.
In this paper, we focus on bridging the generative diffusion models with VLP models to empower the generative diffusion model with the cross-modal discrimination capability, in the spirit of similar principles via large-scale pretraining.

There exist some methods that have considered combining generative models and discriminative models into a single framework. 
The famous generative adversarial networks (GAN)~\cite{goodfellow2020generative} introduce the discriminator to guide the adversarial learning of the generator. 
However, the implicit adversarial of GAN would lead to mode-collapse and training instabilities.
Moreover, HybViT~\cite{Yang2022HibViT} tried to replace the UNet structure in GLIDE~\cite{nichol2021glide} with a ViT~\cite{dosovitskiy2021image} model and directly added a classification head to perform image generation and classification jointly, 
while ignoring the powerful diffusion process for the discriminate task.
On the other hand, recent unified general vision models such as Pix2Seq \cite{chen2021pix2seq}, OFA \cite{wang2022unifying}, and Unified I/O \cite{lu2022unified} try to unify different vision-language tasks into an autoregressive sequence prediction framework. 
However, the autoregressive image generation solutions such as DallE~\cite{ramesh2021zero} and OFA~\cite{wang2022unifying} show inferior performance compared to the DM-based models, such as DallE2~\cite{ramesh2022hierarchical} and Stable Diffusion~\cite{rombach2022high}, in terms of both generation quality and sampling efficiency. 

In this paper, we present \textbf{DiffDis}, 
a unified vision-language diffusion model for both generative and discriminative tasks under the diffusion paradigm.
Specifically, DiffDis first formulates the image-text discriminative problem as a generative diffusion process of the text embedding outputted by the text encoder conditioned on the input image.
Therefore, the generation and discrimination tasks can share the same image-branch network (i.e., original U-Net) in the multi-modality diffusion model. 
During inference, zero-shot image classification is performed by calculating the cosine similarity between the generated text embedding and the downstream text embeddings.
Secondly, we design a dual-stream network architecture to better fuse the knowledge of latent images with different scales into the text query in image-text alignment.
Finally, a unified training paradigm is further proposed to alternatively feed the required inputs and the conditions when jointly performing diffusion-based generative and discriminative tasks.
When training discriminative tasks, the image branch serves as an image encoder to feed conditional information into the reverse text embedding diffusion process and vice versa.

Extensive experiments have shown that the proposed DiffDis method can achieve better performance on both zero-shot classification and text-guided image generation tasks.
Compared to the single-task baseline, our unified framework DiffDis can achieve 1.65\% improvement on the average zero-shot classification accuracy on 12 datasets and a 2.42 improvement on FID of zero-shot image synthesis compared to the single-task model.
This work is the first to unify the training of generative and discriminative tasks under the diffusion process. We hope that this research will serve as an early-stage exploration for future studies aiming to unify these two tasks under the diffusion process, thereby providing more choices for future multi-task multi-modal jointly-training frameworks.

Our contributions can be summarized as:
\begin{itemize}
    \item We propose DiffDis to explore a unified vision-language diffusion model for both multi-modality generation and discrimination tasks.
    \item DiffDis reformulates the image-text discriminative problem by utilizing a generative diffusion process of the text embeddings conditioned on input images.
    \item We propose a dual-stream network architecture and a diffusion-based unified training paradigm for jointly training the generative and discriminative tasks.
    \item Extensive experiments demonstrate that our DiffDis outperforms single-task models, achieving a 1.65\% improvement on average zero-shot classification accuracy across 12 datasets and a 2.42 improvement on FID of text-guided image generation. Additionally, DiffDis outperforms CLIP, with a 4.7\% improvement on average zero-shot classification accuracy across 12 datasets and a 14.5\% improvement on average R@1 of image-text retrieval tasks on Flickr30k and MSCOCO.
\end{itemize}

\section{Related Work}
\begin{figure*}[ht!]
\begin{center}
\includegraphics[width=0.9\linewidth]{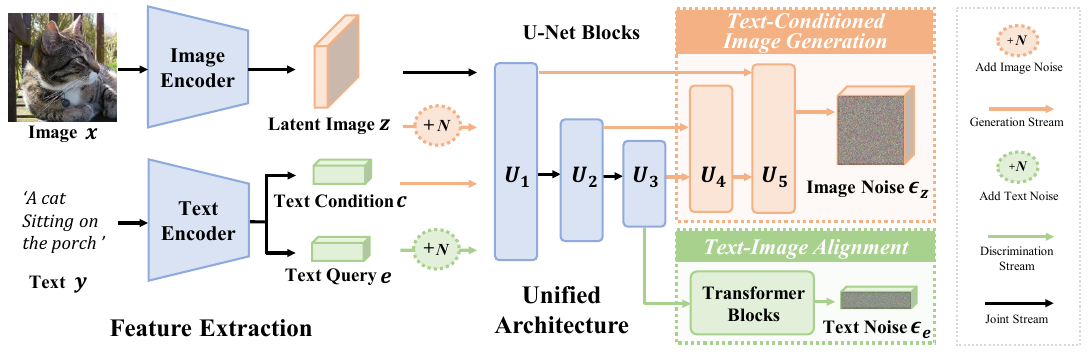}
\end{center}
\vspace{-4mm}
  \caption{Overall model architecture of DiffDis. DiffDis includes an image encoder to encode the image input $x$ into the latent image $z$ and a text encoder to obtain the text condition $c$ and text query $e$ with the caption input $y$. For text-conditional image generation, 
  the latent image $z$ will be added noise and is fed into the UNet with $c$ as the condition to predict the added noise $\epsilon_z$. For image-text alignment learning, the text query $e$ will be added noise and is fed into the UNet with latent image $z$ as the condition to predict the added noise $\epsilon_e$.}
\label{fig:overall_architecture}
\vspace{-3mm}
\end{figure*}

\noindent\textbf{Vision-Language Pre-training.} 
Current communities in natural language processing and computer vision both favor the pre-train-and-fine-tune scheme because of the superior performance of the pre-trained models~\cite{devlin2018bert,brown2020language,dosovitskiy2020image}.
Recent works such as CLIP\cite{radford2021learning} and ALIGN \cite{jia2021scaling} then extend this diagram to a joint cross-modal domain of Vision-and-Language Pre-training (VLP). 
These large-scale models have shown promising results in various downstream tasks, such as zero-shot classification and image-text retrieval.
Their great generalization ability mainly comes from the large-scale automatic-collected image-text dataset from the Internet (e.g, YFCC100M \cite{thomee2016yfcc100m}, CC12M \cite{changpinyo2021cc12m})
The VLP models can be categorized by their pre-training tasks:
(a) Image-text contrastive learning task: CLIP \cite{radford2021learning}, ALIGN \cite{jia2021scaling}, FILIP \cite{yao2022filip} and UNIMO \cite{li2020unimo} utilize the cross-modal contrastive learning which aligns the textual and visual embedding;
(b) Language Modeling (LM) based tasks: VisualBERT \cite{li2019visualbert}, UNITER \cite{chen2020uniter}, M6\cite{lin2021m6}, and DALL-E \cite{ramesh2021zero} employ LM-like objectives, including both masked LM (e.g., Masked Language), and autoregressive LM (e.g., image captioning).
In contrast, we try to follow the diffusion framework to perform image-text alignment learning in an end-to-end unified manner while maintaining the benefit of strong image generation ability.

\noindent\textbf{Denoising Diffusion Probabilistic Models.}~
Since Ho \etal \cite{ddpm} build a connection between diffusion model \cite{diffusion} and denoising score matching model \cite{score-model} and propose DDPMs (Denoising Diffusion Probabilistic Models) to achieve high image generation quality, diffusion models start to attract attention. Nichol \etal \cite{nichol2021improved-ddpm} propose to learn the variances of the reverse diffusion process to achieve higher sampling efficiency with fewer forward passes. Dhariwal \etal \cite{adm} achieve better image generaion quality than GANs by finding a better model architecture through ablations and propose a new sampling technique called classifier guidance. While classifier guidance requires training an extra classifier model, Ho \etal \cite{ho2022classifier-free} propose classifier-free guidance to circumvent this problem by jointly training a conditional and an unconditional model and combine the resulting conditional and unconditional scores to achieve the same effect as classifier guidance. Recently, diffusion models are applied to text-to-image generation and achieve appealing generation results \cite{nichol2021glide,ho2022cascaded-diffusion,ramesh2022hierarchical,imagen,rombach2022high}.
Some methods~\cite{Yang2022HibViT,baranchuk2021label} make an attempt to unify the generation and discriminative tasks by directly adding a classification head into the UNet structure while our DiffDis formulate the discriminative problem into the powerful diffusion process.
\vspace{-2mm}
\section{Methodology}
\vspace{-2mm}

In this section, we first 
review some preliminaries about diffusion models (Sec. \ref{sec:preliminaries}). Then we 
introduce our formulations of generative and discriminate tasks under the unified diffusion process~(Sec. \ref{sec:task_formulation}),
followed by a detailed description of the network architecture of the proposed unified DiffDis~(Sec. \ref{sec:unified_model_architecture}).
Finally, the training paradigm is introduced to deal with the generative and discriminative tasks with different inputs and conditions~(Sec. \ref{sec:training_paradigm}).

\subsection{Preliminary on Diffusion Models}
\label{sec:preliminaries}

Given a sample from the real data distribution $x_0 \sim q(x_0)$, Gaussian diffusion models first produce a Markov chain of latent variables $x_1,...,x_T$ by progressively adding Gaussian noise to the sample according to some variance schedule given by $\beta_t$ as follows:
\begin{equation}
    q\left(x_t \mid x_{t-1}\right) = \mathcal{N}\left(x_t ; \sqrt{1-\beta_t} x_{t-1}, \beta_t I\right),
\end{equation}
and then learn a model 
to approximate the true posterior:
\begin{equation}
        p_\theta\left(x_{t-1} \mid x_t\right) = \mathcal{N}\left(\mu_\theta\left(x_t, t\right), \Sigma_\theta\left(x_t, t\right)\right),
\end{equation}
to perform the reverse denoising process for sampling: starting from a random noise $x_T \sim \mathcal{N}(0, I)$ and gradually reducing the noise to finally get a sample $x_0$. While a tractable variational lower-bound $\mathcal{L}_{VLB}$ on $\log p_\theta(x_0)$ can be used to optimize $\mu_\theta$ and $\Sigma_\theta$, to achieve better results, Ho \etal \cite{ddpm} instead adopt a denoising network $\epsilon_\theta(x_t, t)$ which predicts the noise component of a noisy sample $x_t \sim q(x_t | x_0)$ and the following training objective:
\begin{equation}
    \mathcal{L} = \mathbb{E}_{x_0 \sim q\left(x_0\right), \epsilon \sim \mathcal{N}(0, I), t \sim[1,T]}\left\|\epsilon-\epsilon_\theta\left(x_t, t\right)\right\|^2,
\end{equation}
where $t$ uniformly sampled from $\{1,...,T\}$.

\subsection{Task Reformulation}
\label{sec:task_formulation}

\noindent\textbf{Basic Notations.} We denotes the image-text dataset as $D =\{x_i, y_i\}_{i=1}^N$, where $x_i$ and $y_i$ denote the $i$-th image and text. $N$ is the total number of image-text pairs. As shown in Fig. \ref{fig:overall_architecture}, DiffDis consists of an image encoder $\mathcal{V}$ and text encoder $\mathcal{T}$.
Besides, $\Phi_{u}$ and $\Phi_{\theta}$ represent the UNet model 
used for text-conditional image generation, and the encoder part of UNet model with additional transformer blocks for image-text alignment learning, respectively. 

\noindent\textbf{Diffusion-based Text-conditioned Image Generation.} The generative part of DiffDis aims to generate images conditioned on input text prompts. Following the standard diffusion models~\cite{nichol2021glide,ho2022cascaded-diffusion,ramesh2022hierarchical,imagen,rombach2022high}, DiffDis generates image samples by gradually removing the noise from a random Gaussian noise signal over a finite number of steps. 
The diffusion model is trained by adding and predicting the different levels of noise on the image in the opposite direction of the sampling process.

Specifically, in the training procedure, the diffusion process of image can be represented as a parameterized Markov chain, which adds $T$ steps' random Gaussian noise $\epsilon$ to gradually convert the original image $x_0$ to a random Gaussian distribution $x_T$. 
Following the LDM \cite{rombach2022high}, we utilize the latent image $z = \mathcal{V}(x) \in \mathbb{R}^{H \times W \times d_x}$ outputted by the image encoder $\mathcal{V}$ instead of the RGB space of image $x$ as the input signal.
Therefore, the diffusion-based text-conditional image generation loss $\mathcal{L}_{IG}$, which aims to predict the added Gaussian noise, can be formulated as:
\begin{equation} \label{eq:image-generation-simple-loss}
\mathcal{L}_{IG} = \mathbb{E}_{\mathcal{V}(x), \mathcal{\epsilon}_z \sim \mathcal{N}(0, I), t_z}\left[ \| \epsilon_z - \Phi_{u}(z_t, t_z, c) \|^2 \right]
\end{equation}
where text condition $c = \mathcal{T}(y) \in \mathbb{R}^{L \times d_y}$ denotes the token-wise representations of the text prompt $y$.
$L$ and $d_y$ represent the context length and embedding dimensions of outputted token embeddings, respectively.

During the inference~(sampling) process, starting from a random Gaussian noise $z_T \sim \mathcal{N}(0, I)$, we reverse the diffusion process and gradually remove the predicted noise $\Phi_{u}(z_t, t_z, c)$ to obtain the sampled latent image after finite steps. We use an image decoder $\mathcal{D}$ to convert the  sampled latent image back to RGB space $\hat{x} = \mathcal{D}(\hat{z}_0)$. DDIM sampler \cite{song2020denoising} and classifier-free guidance \cite{ho2022classifier-free} are employed. More details can refer to Algorithm \ref{alg:image generation sampling}. 


\noindent\textbf{Diffusion-based Image-text Alignment Pretraining.} Previous methods perform image-text alignment pretraining by aligning the visual feature and textual feature in a common semantic space, where the positive image-text pairs are pulled towards each other and the negative image-text pairs are pushed against each other~\cite{yao2022filip, jia2021scaling}. The image-text contrastive~(ITC) loss, like in CLIP~\cite{radford2021learning}, first calculates the cosine similarity of normalized visual feature $v_i$ and textual feature $e_j$ to measure the relevance of each image and text: $s^y_{i,j} = s^x_{i,j} = v_i^\top e_j,$
where $s^x_{i,j}$, $s^y_{i,j}$ denote the image-to-text similarity and the text-to-image similarity. Besides, $v_i$ is the global representation of the image $x_i$ and $e_j$ is the global representation of the text $y_i$. 
Based on $s^x_{i,j}$ and $s^y_{i,j}$, the ITC loss can be calculated as \cite{radford2021learning}.


To reformulate this image-text alignment problem into a diffusion process, 
we propose the denoising diffusion-based image-text alignment which treats the latent image $z$ as the condition and learns the distribution of the corresponding text embedding $e \in \mathbb{R}^{1 \times d_y}$. The diffusion process of $e$ can be formulated as 
\begin{align}
    e_t &= \gamma \sqrt{\bar{\alpha}_t} e_0 + \sqrt{1-\bar{\alpha}_t} \epsilon_e, \\
    \alpha_t &= 1-\beta_t, \bar{\alpha}_t = \prod_{j=0}^t \alpha_j,
\end{align}
where we omit the index $i$, and $\beta_t$ is used to control the strength of added noise for timestep $t_e$. The $\gamma \in \left(0, 1\right]$ is the scale factor to scale the text embedding $e_0$.

In contrast to the image generation task, the diffusion model $\Phi_\theta$ is trained to estimate the original clean text query $\hat{e}_0=\Phi_\theta(e_t, t_e, z)$. Note that $\Phi_\theta$ can also be a noise prediction model to predict the noise $\epsilon_e$.


Then the diffusion-based image-text alignment objective is to minimize the distance between $\hat{e}_0$ and $e$.
In detail, we first calculate the cosine similarity between $\hat{e}_0$ and $e$:
\begin{align}
    s =\hat{e}_0 ^\top e, \label{eq:modified_similarity}
\end{align}
Note that we omit the index of the embeddings here.
Then the diffusion-based image-text alignment loss $\mathcal{L}_{ITA}$ can be calculated as:
\begin{align}
\small
\mathcal{L}_{ITA} =-\frac{1}{2B}\sum^{B}_{i=1}[\log \frac{\exp (s_{i,i}^x)}{ \sum_{j} \exp (s_{i,j}^x)} +
\log \frac{\exp (s_{i,i}^y)}{ \sum_{j} \exp (s_{j, i}^y)}], \label{eq:new_contrastive_loss}
\end{align}
where $B$ denotes the batch size and $\mathcal{L}_{ITA}$ is calculated for each diffusion step during training.


\begin{figure}[t!]
\begin{center}
\includegraphics[width=0.92\linewidth]{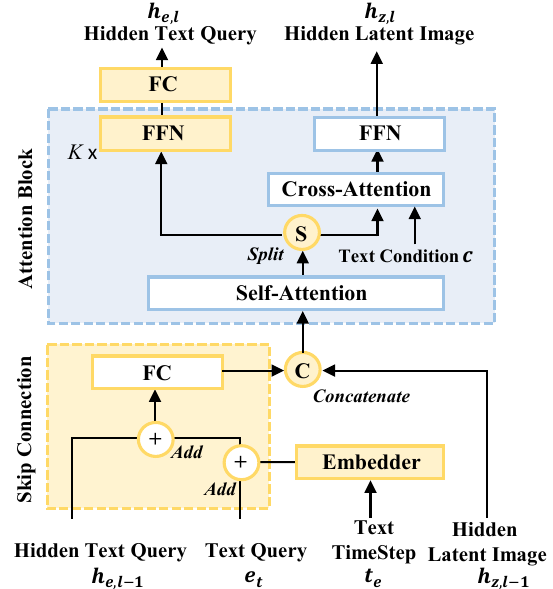}
\end{center}
  \caption{Detailed architecture of dual-stream deep fusion attention block. Compare to the conventional attention block (blue part) in \cite{rombach2022high}, we design a dual-stream deep fusion attention architecture to better fuse the knowledge of latent image into the text query in cross-modal alignment learning. The separate FFNs learn modality-specific information. Besides, we build a cross-block skip connection from the noised text query $e_t$ with time condition to the hidden text query $h_{e,l-1}$ outputted from the last block.}
\label{fig:detailed arch}
\end{figure}

\subsection{Unified Model Architecture}
\label{sec:unified_model_architecture}

\noindent \textbf{Feature Extraction Modules.} As shown in Fig. \ref{fig:overall_architecture}, DiffDis includes an image encoder to obtain the latent image \cite{rombach2022high} and a text encoder to obtain the text condition and text query.
Specifically, the image encoder $\mathcal{V}$ of an autoencoder aims to convert the image $x$ from RGB space to image's latent representation $z$ which improves the training efficiency and perform image generation on high-resolution image synthesis. 
On the other hand, we adopt the text encoder $\mathcal{T}$ to encode the text prompt $y$ to the text condition $c$ and text query $e$. Note that the dimensions of these two text embedding outputs are different. 
The text condition $c \in \mathbb{R}^{L \times d_y}$ is the token-wise representation of the text prompt $y$ while the text query $e \in \mathbb{R}^{1 \times d_y}$ is the normalized global representation of the text prompt $y$.


\noindent\textbf{Unified Architecture.} The following UNet's structure receives three inputs: latent image $z$, text condition $c$ and text query $e$. 
For \textit{text-conditional image generation task}, we adopt the noisy latent image $z_t$ and text condition $c$ as input to predict the noise $\epsilon_z$ added on latent image $z$.
For \textit{diffusion-based image-text alignment learning}, we adopt the latent image $z$ and noisy text query $e_t$ as input to predict the original clean text query $e$. Note that it is viable to predict the noise $\epsilon_e$ added on text query $e$ for the alignment learning.
More training details can refer to Sec.~\ref{sec:training_paradigm}.
Besides, the $\Phi_{\theta}$ also contains a unique transformer with $M$ transformer block and a linear predictor. The unique transformer follows the middle block of the UNet to obtain more semantic information.
The transformer's input is the concatenation of the flattened image's feature map and the text query outputted from the middle block of UNet.
Finally, the linear predictor will be fed by the text query token and predict the original clean text query $e$.



\noindent\textbf{Dual-stream Deep
Fusion Attention Block.} We modify the architecture of attention blocks in UNet to better unify these two tasks with different inputs. 
Previous stable diffusion \cite{rombach2022high} directly use $K$ transformer decoder blocks to inject the text condition information into the latent image via the cross-attention mechanism.
As illustrated in Fig. \ref{fig:detailed arch}, we proposed a dual-stream deep fusion block to better fuse the latent image knowledge into the text query for cross-modal alignment learning.
Specially, we adopt an additional fully-connected layer to project the text query embedding into the same dimension as the hidden latent image space. 
Then the concatenation of the projected text query and the hidden latent image $h_{z,l-1}$ outputted by the last block will pass $K$ transformer blocks. In each transformer block, we propose the modality-specific feedforward neural network (FFN) for text and image. Besides, the text query will skip the cross-attention layer.
Following the transformer blocks, we separate the concatenation between the text query and the image hidden, and project the text query back to the text embedding space using a fully-connected layer.
Furthermore, we build a cross-block skip connection from the noised text query $e_t$ with time condition to the hidden text query $h_{e,l-1}$ outputted by the last block. The hidden text query $h_{e,l-1}$ is initialized to zero.


\begin{algorithm}[t!]
\caption{Diffusion-base Unified Training}
\label{alg:training paradigm algorithm}
\textbf{Input:} Image $x$, Text $y$, Image Encoder $ \mathcal{V}$, Text Encoder $ \mathcal{T}$, Timestep $T$.
\begin{algorithmic}[1]
\Repeat
    \State $c, e = \mathcal{T}(y)$ \Comment{Get text condition and text query}
    \State $z = \mathcal{V}(x)$    
    \State $t_z, t_e \sim \text{Uniform}(\{1, \dots T\})$
    \State $\epsilon_z, \epsilon_e \sim \mathcal{N}(0,I)$
    \State Mask $e$ and calculate $\mathcal{L}_{IG}$ based on Eq. \ref{eq:image-generation-simple-loss}.
    \State Mask $c$ and calculate $\mathcal{L}_{ITA}$ based on Eq. \ref{eq:new_contrastive_loss}.
    \State Calculate total loss $\mathcal{L}$ based on Eq. \ref{eq:total-loss}
    \State Take gradient descent step on
    \Statex \qquad \qquad \qquad$\nabla_{u, \theta}~ \mathcal{L}_{Total}$
\Until{converged}
\end{algorithmic}
\end{algorithm}

\subsection{Diffusion-base Unified Training}
\label{sec:training_paradigm}
In this section, we introduce the training paradigm to unify diffusion-based image generation training and cross-modal alignment learning into a single framework.  Algorithm \ref{alg:training paradigm algorithm} illustrates the whole training algorithm. 

To alternatively feed the required inputs and the conditions when performing diffusion-based generation and alignment tasks, we introduce the masking mechanism to erase the unnecessary input.
Note that the whole inputs' space includes the latent image $z$ outputted by the image encoder, text condition $c$ and text query $e$ outputted by the text encoder.
For the \textit{image generation task}, we mask the text query $e$, and feed the noisy latent image $z_t$ with time step $t$ and text condition $c$ into the following UNet model.
While for \textit{image-text alignment learning}, to avoid leaking textual information,
we mask the text condition $c$, then feed the noisy text query $e_t$ with timestep $t$ and the latent image $z$ as condition.
Considering the image generation loss $\mathcal{L}_{IG}$ from Eq. \ref{eq:image-generation-simple-loss} and the diffusion-based image-text alignment loss $\mathcal{L}_{ITA}$ from Eq. \ref{eq:new_contrastive_loss}, the total loss of DiffDis $\mathcal{L}_{Total}$ can be calculated as:
\begin{equation}
    \mathcal{L}_{Total} = \mathcal{L}_{IG} + \lambda \mathcal{L}_{ITA}
    \label{eq:total-loss}
\end{equation}
where $\lambda$ denotes the weight factor.

Algorithm \ref{alg:image generation sampling} and Algorithm \ref{alg:image-tet contrastive inference} show the inference algorithm of sampling processes of text-conditional image generation and image-text alignment, respectively. During inference, different sampling acceleration techniques, such as classifier-free guidance sampling \cite{ho2022classifier-free}, can also be seamlessly integrated into our framework for both tasks.

\begin{algorithm}[t]
\caption{Text-conditional Image Generation Sampling.}
\label{alg:image generation sampling}
\textbf{Input:} Text $y$, Text Encoder $ \mathcal{T}$, Image Decoder $\mathcal{D}$, Timestep $T$, Noise Schedule $\{\beta_t\}_{t=1}^T$, Classifier-free guidance scale $w$.
\begin{algorithmic}[1]
\State $z_T \sim \mathcal{N}(0, I)$
\State $c = \mathcal{T}(y)$
\State $\alpha_t = 1-\beta_t, \bar{\alpha}_t = \prod_{k=1}^t \alpha_t$


\For{$t=T,\dots,1$}  \Comment{For simplify, $t$ stands for $t_z$.}
    \State $\hat{\epsilon}_z = (1+w)\Phi_u(z_t, t, c) - w \Phi_u(z_t, t)$
    
    \State $z_{t-1}= \sqrt{\bar{\alpha}_{t-1}}(\frac{z_t - \sqrt{1-\bar{\alpha}_t} \hat{\epsilon}_z}{\sqrt{\bar{\alpha}_t}}) + \sqrt{1-\bar{\alpha}_{t-1}} \hat{\epsilon}_z
    $
\EndFor
\State \Return $\mathcal{D}(z_0)$
\end{algorithmic}
\end{algorithm}

\begin{algorithm}[t]
\caption{Image-text Alignment Inference.}
\label{alg:image-tet contrastive inference}
\textbf{Input:} Image $x$, Text $y'$ of Downstream Task, Image Encoder $ \mathcal{V}$, Text Encoder $ \mathcal{T}$, Timestep $T$, Noise Schedule $\{\beta_t\}_{t=1}^T$, Classifier-free guidance scale $w$.
\begin{algorithmic}[1]
\State $e_T \sim \mathcal{N}(0, I)$
\State $z = \mathcal{V}(x)$
\State $\alpha_t = 1-\beta_t, \bar{\alpha}_t = \prod_{k=1}^t \alpha_t$


\For{$t=T,\dots,1$}  \Comment{For simplify, $t$ stands for $t_e$.}
    \State $\hat{e}_0 = (1+w)\Phi_\theta(e_t, t, z) - w \Phi_\theta(e_t, t)$
    
    \State $\hat{\epsilon}_e = \left(e_t - \sqrt{\bar{\alpha}_t} \hat{e}_0\right) / \sqrt{1-\bar{\alpha}_t}$

    \State $e_{t-1}= \sqrt{\bar{\alpha}_{t-1}}(\frac{e_t - \sqrt{1-\bar{\alpha}_t} \hat{\epsilon}_e}{\sqrt{\bar{\alpha}_t}}) + \sqrt{1-\bar{\alpha}_{t-1}} \hat{\epsilon}_e$

\EndFor

\State $e_0 = e_0 / \|e_0\|$ \Comment{L2 Normalize}
\State $e = \mathcal{T}(y')$   \Comment{Extract Text Embedding}
\State $e = e / \|e\|$ \Comment{L2 Normalize}
\State Perform the similarity $e_0^\top e$ on downstream tasks.
\end{algorithmic}
\end{algorithm}

\begin{table*}[t]
\small
\setlength{\tabcolsep}{6.5pt}
\renewcommand\arraystretch{0.92}
\begin{center}
\begin{tabular}{l|cccc cccc cccc|c}
 &
\rotatebox{90}{CIFAR10}~~ &
\rotatebox{90}{CIFAR100}~~ &
\rotatebox{90}{Caltech101}~~ &
\rotatebox{90}{StanfordCars}~~ &
\rotatebox{90}{Flowers102}~~ &
\rotatebox{90}{Food101}~~ &
\rotatebox{90}{SUN397}~~ &
\rotatebox{90}{DTD}~ &
\rotatebox{90}{Aircrafts}~~ &
\rotatebox{90}{OxfordPets}~~ &
\rotatebox{90}{EuroSAT}~~ & 
\rotatebox{90}{\textbf{ImageNet}}~~ &
\rotatebox{90}{\textbf{Average}}~~ \\
\midrule
CLIP-ViT-B/32 & 
\textbf{63.0} & 
28.5 & 
58.2 & 
1.1 & 
12.4 & 
12.8 & 
24.3 & 
7.6 & 
1.5 & 
11.5 & 
11.8 & 
16.7 & 
20.8  \\
CLIP-ViT-B/16 & 
57.2 &
27.0 &
54.8 &
1.0 &
12.9 &
13.6 &
28.8 &
10.1 &
1.1 &
10.3 &
10.8 &
19.7 &
20.6  \\
CLIP-ViT-L/14 &
57.5 &
28.1 &
57.2 &
1.8 &
10.8 &
14.3 &
31.2 &
12.1 &
1.7 &
11.7 &
\textbf{24.9} &
21.1 &
22.7  \\
DiffDis &
57.1 &
\textbf{32.3} &
\textbf{68.6} &
\textbf{3.1} &
\textbf{16.9} &
\textbf{16.2} &
\textbf{35.1} &
\textbf{26.4} &
\textbf{2.0} &
\textbf{28.5} &
17.5 &
\textbf{25.9} &
\textbf{27.4} \\
\bottomrule
\end{tabular}
\end{center}
\caption{Top-1 accuracy(\%) of zero-shot image classification on 12 datasets. Note that CLIP models are pre-trained on CC3M by using 4x larger batch size and 3.3x longer training epochs.
}
\label{zeroshot-classification-table}
\end{table*}

\section{Experiments}

In this section, we first describe the detailed experiment settings (Sec. \ref{sec:Experiment Setting and Implementation Details}). Then we show the results on zero-shot image classification, image-text retrieval 
and text-to-image generation (Sec. \ref{sec:Zero-shot Image Classification}). 
Finally, we conduct ablation studies on our DiffDis to validate the effectiveness of implementation designs (Sec. \ref{sec:Ablation Study}).

\subsection{Experiment Setting}\label{sec:Experiment Setting and Implementation Details}
\noindent\textbf{Model Architecture.}~~To obtain better image synthesis performance, we initialize the autoencoder and UNet from Stable Diffusion-v1-1\footnote{Stable-Diffusion-v1-1 checkpoint: https://huggingface.co/CompVis/stable-diffusion-v-1-1-original}. The transformer in model $\Phi_{\theta}$ is trained from scratch and contains 6 transformer blocks with 768 model width and 64-dim attention heads. The text encoder is initialized from CLIP-ViT-L/14 \cite{radford2021learning}.

\begin{table*}
\center
\setlength{\tabcolsep}{5pt}
\renewcommand\arraystretch{0.92}
\small
\begin{tabular}{l|cccccc | cccccc|c}
\toprule
\multirow{3}{*}{Model} & 
\multicolumn{6}{c|}{Flickr30K}   & \multicolumn{6}{c|}{MSCOCO}                                                                  \\
    & \multicolumn{3}{c}{image-to-text} & \multicolumn{3}{c|}{text-to-image} & \multicolumn{3}{c}{image-to-text} & \multicolumn{3}{c|}{text-to-image} \\
    & R@1           & R@5           & R@10         & R@1           & R@5           & R@10         & R@1           & R@5           & R@10         & R@1           & R@5           & R@10  & Mean R@1       \\
\midrule
CLIP-ViT-B/32 &   18.8 &  42.3 &  53.9 &  12.5 &   30.5 &  39.9 & 10.1 & 25.3 &  35.3 &  7.0 &  18.9 &  27.1 & 12.1 \\
CLIP-ViT-B/16 &  30.1 &	 56.4 &	68.9 & 19.4 &  42.1  & 52.7 &  14.4 &  34.6 & 46.1 & 10.3 & 26.5 & 36.6 & 18.6 \\
CLIP-ViT-L/14 &  29.9 &  58.3 &	70.4 &  20.3 & 46.1 & 57.6 & 14.7 & 35.0 & 47.1 &  11.3 & 28.4 & 38.9 & 19.1 \\
DiffDis       &  \textbf{49.8}  & \textbf{77.5} &  \textbf{85.6}  & \textbf{38.8} & \textbf{67.9}  & \textbf{77.6} &  \textbf{26.3} &  \textbf{50.9}  & \textbf{62.8} &  \textbf{19.5} & \textbf{42.2} & \textbf{54.2} & \textbf{33.6} \\
\bottomrule         
\end{tabular}
\vspace{2mm}
\caption{Results of zero-shot image-text retrieval on Flickr30K and MSCOCO datasets. `R@K' means top-K recall. `Mean R@1' means the average R@1 of image-to-text retrieval and text-to-image retrieval on Flickr30K and MSCOCO. Note that CLIP models are pre-trained on CC3M by using 4x larger batch size and 3.3x longer training epochs.}
\label{tab:zero-shot-retrieval-table}
\end{table*}


\noindent\textbf{Experiment Details.}~~We pre-train models on Conceptual Caption Dataset (CC3M) \cite{sharma2018conceptual} to evaluate our DiffDis's effectiveness. The resolutions of the original image and the latent image are set as $256 \times 256 \times 3$ and $32 \times 32 \times 4$ respectively. 
Following Stable Diffusion, the pixel values of the image are normalized to [-1, 1] and the autoencoder and text encoder are frozen.
We pre-train our DiffDis model for 6 epochs using the AdamW \cite{loshchilov2018decoupled} optimizer with weight decay of 1e-4. The batch size is set as 256 and the learning rate is set as 1e-5 with 1000 steps linear warmup and kept unchanged until the training is finished. The new parameters introduced for the discriminative tasks use the learning rate of 1e-4. We simply set $\lambda$ to 1 and randomly drop 10\% text condition for image generation and 10\% image condition by zeroing the image for diffusion-based image-text alignment to enable the classifier-free guidance \cite{ho2022classifier-free}. 
Exponential moving average (EMA) is applied every iteration and the decay coefficient is set as 0.9999. The EMA model is utilized to evaluate the performance of downstream tasks.

\noindent\textbf{Evaluations.} 
For \textit{zero-shot image classification}, we evaluate our proposed DiffDis model on 12 classification datasets, i.e.,  CIFAR10 \cite{krizhevsky2009learning}, CIFAR100 \cite{krizhevsky2009learning}, Caltech101 \cite{fei2006one}, StanfordCars~\cite{krause20133d}, Flowers102~\cite{nilsback2008automated}, Food101 \cite{bossard2014food}, SUN39 \cite{barriuso2012notes}, Describable Textures Dataset (DTD) \cite{cimpoi2014describing}, Aircrafts~\cite{maji2013fine},  OxfordPets \cite{parkhi2012cats}, EuroSAT~\cite{helber2019eurosat}, and ImageNet~\cite{deng2009large}. We adopt DDIM sampler \cite{song2020denoising} as described in Algorithm \ref{alg:image-tet contrastive inference} to predict the text embedding and calculate the similarity between predicted text embedding and text embedding. During inference, we use 8 sampling steps and classifier-free guidance scale~\cite{ho2022classifier-free} of 3. Following CLIP~\cite{radford2021learning}, we ensemble the prompt templates to improve the zero-shot classification performance by averaging the text embeddings across different prompt templates. 

For \textit{zero-shot image-to-text retrieval} and \textit{text-to-image retrieval tasks}, we conduct experiments on karpathy split test set~\cite{karpathy2015deep} of MSCOCO~\cite{lin2014microsoft} and Flickr30K~\cite{plummer2015flickr30k} which are widely used benchmark datasets. The inference processing of retrieval is similar to image classification.

For \textit{zero-shot text-to-image generation}, we evaluate the  performance on 30,000 text prompts from MSCOCO dataset~\cite{lin2014microsoft} under the evaluation of FID, KID by applying torch-fidelity~\cite{obukhov2020torchfidelity}. We also calculate CLIP-Score~\cite{hessel2021clipscore} under pre-trained CLIP-RN50. We compare our proposed DiffDis with Stable Diffusion fine-tuned on CC3M with the same training hyper-parameters. All models use PNDM sampler~\cite{liu2022pseudo} with 50 sampling steps under the classifier-free guidance scale~\cite{ho2022classifier-free} of 3. 

\subsection{Main Results}\label{sec:Zero-shot Image Classification}

\noindent\textbf{Zero-shot Image Classification.} 
Table \ref{zeroshot-classification-table} presents detailed results on 12 classification datasets, comparing CLIP-ViT-B/32, CLIP-ViT-B/16, and CLIP-ViT-L/14 pre-trained on the same dataset, i.e., CC3M. To ensure a fair comparison with DiffDis, all CLIP models' text encoders are initialized from the pre-training~\cite{radford2021learning}. Despite CLIP models being pre-trained with a 4x larger batch size and 3.3x longer training epochs, DiffDis achieves an average accuracy gain of 4.7\% across 12 datasets, outperforming CLIP-ViT-L/14. Moreover, DiffDis demonstrates significant performance improvements on some domain-specific datasets such as OxfordPets, which we attribute to its strong generation ability to capture fine-grained information.



\noindent\textbf{Zero-shot Image-Text Retrieval.} 
Table \ref{tab:zero-shot-retrieval-table} presents experimental results on image-to-text (I2T) retrieval and text-to-image (T2I) retrieval for Flickr30K and MSCOCO datasets. We can observe that DiffDis outperforms CLIP-ViT-L/14 in R@1 of I2T retrieval and T2I retrieval by 11.6\% and 8.2\%, respectively, on the MSCOCO dataset. Moreover, on the Flickr30K dataset, DiffDis achieves superior results with 19.9\% and 18.5\% improvements in R@1 of I2T retrieval and T2I retrieval, respectively.


\begin{table}
\center
\setlength{\tabcolsep}{5pt}
\renewcommand\arraystretch{0.92}
\small
\begin{tabular}{l ccc}
\toprule
\multirow{2}{*}{Model}  & 
\multicolumn{3}{c}{\multirow{1}{*}{Text-to-Image Generation)}} \\   
    &  FID$\downarrow$     &  KID$\downarrow$ & CLIP-Score$\uparrow$  \\
\midrule
OFA (Finetuned)~\cite{wang2022ofa}    & 10.5  & -  & -  \\
LAFITE~\cite{zhou2021lafite} & 26.9 & -   & -\\
DALLE~\cite{ramesh2021zero} & 17.8  & -     & -\\
GLIDE~\cite{nichol2021glide} & 12.2  & -    & -\\
DALLE2~\cite{ramesh2022hierarchical} & 10.4  & -   & - \\
Stable Diffusion    & 10.8    & 2.9e-3  &\textbf{24.6}    \\
\textbf{DiffDis}    & \textbf{9.8} & \textbf{2.3e-3} & 24.4 \\
\bottomrule         
\end{tabular} 
\vspace{2mm}
\caption{Quantitative evaluation of FID and CLIP-score on MSCOCO dataset for zero-shot text-guided 256 $\times$ 256 image synthesis. The Stable Diffusion and DiffDis are pre-trained on CC3M.}
\label{tab:image-generation-coco-table}
\end{table}

\begin{table}
\center
\setlength{\tabcolsep}{5pt}
\renewcommand\arraystretch{0.92}
\begin{tabular}{lccc}
\toprule
 Task & FID$\downarrow$ & ZS-Acc$\uparrow$ & Mean R@1$\uparrow$ \\
\midrule
 Dis   &  --   & 11.31 & 15.82 \\
 Gen   & 43.47 & -- & -- \\
Hybrid& \textbf{41.05} & \textbf{12.96} & \textbf{19.72} \\
\bottomrule         
\end{tabular}
\vspace{2mm}
\caption{Results of DiffDis with train-from-scratch UNet.}
\label{tab:comparison-no-pretrain-table}
\end{table}

\begin{figure}
\begin{center}
\includegraphics[width=1\linewidth]{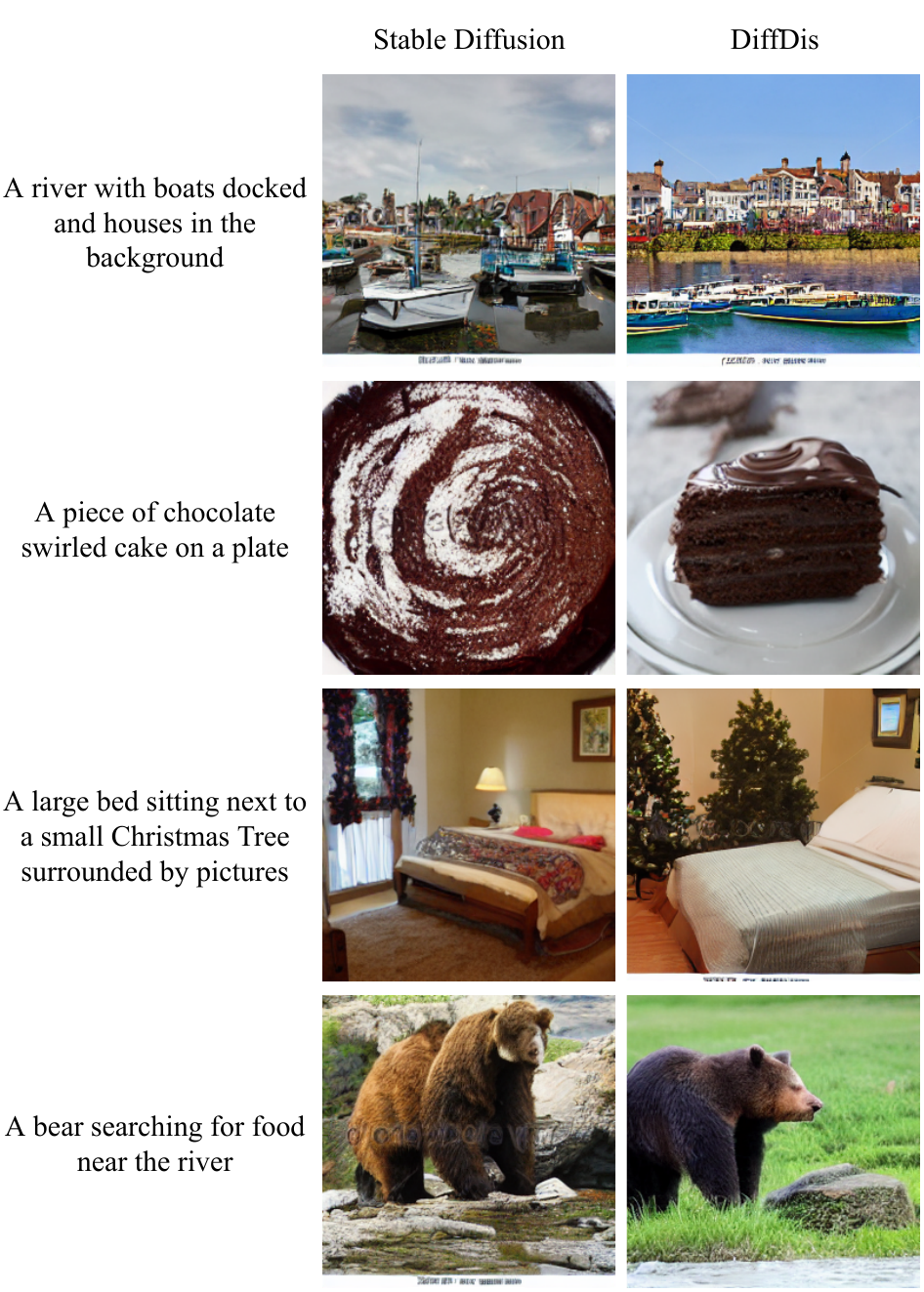}
\end{center}
  \caption{Qualitative comparisons of Stable diffusion and our DiffDis on MSCOCO zero-shot text-to-image generation.}
\label{fig:image_generation_samples}
\end{figure}

\noindent\textbf{Zero-shot Text-to-Image Generation.} 
In addition to the Stable Diffusion model pre-trained on CC3M, we compare our proposed DiffDis model with LAFITE~\cite{zhou2021lafite}, DALLE~\cite{ramesh2021zero}, GLIDE~\cite{nichol2021glide}, DALLE2~\cite{ramesh2022hierarchical}, and OFA~\cite{wang2022ofa}. Table \ref{tab:image-generation-coco-table} demonstrates that our proposed DiffDis model makes a comparable performance to Stable Diffusion, achieving a 1.0 improvement in FID and similarly CLIP-Score, indicating that dual-task learning helps in image generation tasks. Additionally, DiffDis's zero-shot image generation performance on the MSCOCO dataset surpasses OFA, which has been fine-tuned on the MSCOCO dataset. 
Note that OFA is also a hybrid model to combine discriminative tasks and generative tasks. Figure \ref{fig:image_generation_samples} illustrates the synthetic samples under the text condition generated by different models on MSCOCO. 

\subsection{Ablation Study}
\label{sec:Ablation Study}


\noindent\textbf{The Effect of Dual-task Learning.}
Table \ref{tab:comparison-no-pretrain-table} presents a comparison between dual-task learning and single-task learning. In this experiment, UNet is trained from scratch, and the text encoder is frozen, which is the same as stable diffusion pre-training. Compared to single-task learning, dual-task learning shows a 1.65\% improvement in zero-shot ImageNet classification, a 3.9\% improvement on average R@1 of Flickr30k and MSCOCO, and a 2.42 improvement in FID of zero-shot MSCOCO text-to-image generation. These results demonstrate the effectiveness of unifying cross-modal generative and discriminative pre-training into a single framework under the diffusion process.



\noindent\textbf{The Effect of Freezing Text Encoder.} The generative task usually applies a frozen text encoder but the discriminative task often trains the text encoder. Table \ref{tab:effect-of-train-from-scratch-param-table} demonstrates that freezing the text encoder obtains better performance on all tasks, especially in the image generation task. We freeze the text encoder in our main results.

\noindent\textbf{Enlarging the Learning Rate of Train-From-Scratch Parameters.} During pre-training, we utilize a learning rate for the train-from-scratch parameters, i.e., the additional parameters for discriminative tasks, that is 10 times larger than the base learning rate. Specifically, we set the learning rate of 1e-4 for train-from-scratch parameters and the pre-trained parameters use a learning rate of 1e-5. Table \ref{tab:effect-of-train-from-scratch-param-table} demonstrates that enlarging the learning rate of train-from-scratch parameters can improve the performance on three downstream tasks.

\begin{table}
\center
\setlength{\tabcolsep}{5pt}
\renewcommand\arraystretch{0.92}
\resizebox{\linewidth}{!}{
\begin{tabular}{ccccc}
\toprule
Freeze $\mathcal{T}$ & Enlarge LR &  FID$\downarrow$ & ZS-Acc$\uparrow$ & Mean R@1$\uparrow$ \\
\midrule
\xmark & \cmark & 11.56 & 25.72 & 29.37 \\
\cmark & \xmark & 10.98 & 24.98 & 31.08 \\
\cmark & \cmark  & \textbf{9.80} & \textbf{25.92} & \textbf{33.60} \\
\bottomrule         
\end{tabular}
}
\vspace{2mm}
\caption{The effect of enlarging the learning rate for train-from-scratch parameters and freeze the text encoder $\mathcal{T}$.}
\label{tab:effect-of-train-from-scratch-param-table}
\end{table}

\noindent\textbf{The Effect of Text Embedding Scale Factor $\gamma$.}
The experimental results presented in Table \ref{tab:comparison-different-scale-table} examine the impact of text embedding scale factor $\gamma$ for discriminative learning on three downstream tasks. The results show that as $\gamma$ decreases, the performance of text-guided image generation and zero-shot ImageNet classification improves.


\noindent\textbf{The Effect of Classifier-free Guidance.}
Table \ref{tab:different-guidance-table} demonstrates that classifier-free guidance can enhance both image generation and discriminative ability. Specifically, the results indicate that the use of classifier-free guidance with a value of 3 leads to the best performance in terms of image synthesis and zero-shot ImageNet classification.


\begin{table}
\center
\setlength{\tabcolsep}{5pt}
\renewcommand\arraystretch{0.92}
\begin{tabular}{lccc}
\toprule
 $\gamma$ & FID$\downarrow$ & ZS-Acc$\uparrow$ & Mean R@1$\uparrow$ \\
\midrule
 1    & 11.90 & 22.35 & 28.59
 \\
 0.1  & 11.62 & 22.97 & \textbf{29.08}
 \\
 0.01 & \textbf{11.52} & \textbf{23.45} & 28.64
 \\
\bottomrule         
\end{tabular}
\vspace{2mm}
\caption{Results of DiffDis with different scale factor $\gamma$ for text embedding in diffusion-based image-text alignment. The model is trained without enlarging the learning rate, freeze $\mathcal{T}$, and deep fusion blocks.}
\label{tab:comparison-different-scale-table}
\end{table}

\begin{table}
\center
\setlength{\tabcolsep}{5pt}
\renewcommand\arraystretch{0.92}
\resizebox{\linewidth}{!}{
\begin{tabular}{lcccccc}
\toprule
Classifier-free Guidance Scale & None & 2 & 3 & 4 & 5  \\
\midrule
ZS-Acc (ImageNet)$\uparrow$  &  23.15 & 
           23.44 & 
           \textbf{23.45} & 
           23.43 & 
           23.37 \\
FID (MSCOCO)$\downarrow$  &  30.13 &
           12.59 &
           \textbf{11.52} &
           13.25 &
           15.18 \\
\bottomrule         
\end{tabular}
}
\vspace{2mm}
\caption{The ablation of classifier-free guidance level \cite{ho2022classifier-free}. By reformulating in a diffusion framework, Classifier-free guidance can be applied to boost both generative and discriminative tasks.}
\label{tab:different-guidance-table}
\end{table}

\noindent\textbf{The Effect of Different Sampling Steps.} 
In image generation tasks, larger steps for denoising typically result in better performance. This section aims to analyze the impact of various generation steps on the discriminative task. The experimental results are presented in Table \ref{tab:effect-of-different-steps-table}. The results indicate that using 8 steps leads to the best performance on zero-shot ImageNet classification task. However, increasing the number of steps beyond 8 results in longer inference times without a significant improvement in performance.


\begin{table}
\center
\setlength{\tabcolsep}{5pt}
\renewcommand\arraystretch{0.92}
\begin{tabular}{lccccccc}
\toprule
Steps & 1 & 4 & 8 & 10 & 20 & 50  \\
\midrule
ZS-Acc &  14.85 &
23.14 &
\textbf{23.15} &
23.12 &
23.13 &
23.13 \\
\bottomrule         
\end{tabular}
\vspace{2mm}
\caption{The performance on zero-shot ImageNet classification with different sampling steps. No classifier-free guidance.}
\vspace{-3mm}
\label{tab:effect-of-different-steps-table}
\end{table}

\noindent\textbf{Stochasticity of the Discriminative Evaluation.} We run 10 times evaluations of zero-shot classification with different random seeds on five classification datasets to calculate the mean and variance of the performance. Table~\ref{tab:stochasticity-of-zeroshot-cls-table} shows that the discriminative performance is stabled.

\begin{table}
\center
\setlength{\tabcolsep}{5pt}
\renewcommand\arraystretch{0.92}
\Large
\resizebox{\linewidth}{!}{
\begin{tabular}{lccccc}
\toprule
Dataset  & 
\rotatebox{45}{CIFAR10}~~ &
\rotatebox{45}{CIFAR100}~~ &
\rotatebox{45}{Food101}~~ &
\rotatebox{45}{SUN397}~~ &
\rotatebox{45}{\textbf{ImageNet}}~~ 
\\
\midrule
Mean & 57.15 &
32.31 &
16.25 &
35.15 &
25.92  \\
Var & 3.99e-05  &
1.66e-05  &
6.27e-06  &
8.88e-05  &
9.20e-05   \\
\bottomrule         
\end{tabular}
}
\vspace{2mm}
\caption{The stochasticity of DiffDis on zero-shot classification.}
\vspace{-3mm}
\label{tab:stochasticity-of-zeroshot-cls-table}
\end{table}

\section{Conclusion}
\label{sec:Conclusion}
This paper proposes a novel framework named DiffDis, which unifies the training of generative and discriminative tasks under the diffusion process. Initially, we formulate the image-text alignment into a diffusion process by adopting the text embeddings as diffusion objectives. We then propose a dual-stream network architecture that can simultaneously learn to reconstruct the image and text embeddings.
As the first work to unify the training of generative and discriminative tasks under the diffusion process, we conducted a careful ablation study of the different key settings of the unified model. We observed that the unified model achieves improvement in both discriminative tasks and generative tasks compared to the single-task model. We hope that our work can serve as an exploration for future research on unifying discriminative tasks under the diffusion process, thereby providing more choices for future multi-task multi-modal joint-training model frameworks. 

\section{Acknowledgments}
We gratefully acknowledge the support of MindSpore\footnote{\url{https://www.mindspore.cn/}}, CANN~(Compute Architecture for Neural Networks) and Ascend AI Processor used for this research.

{\small
\bibliographystyle{ieee_fullname}
\bibliography{main_review}
}

\clearpage
\appendix

\section{Failure Case}

\begin{figure}[h!]
\begin{center}
\includegraphics[width=1.0\linewidth]{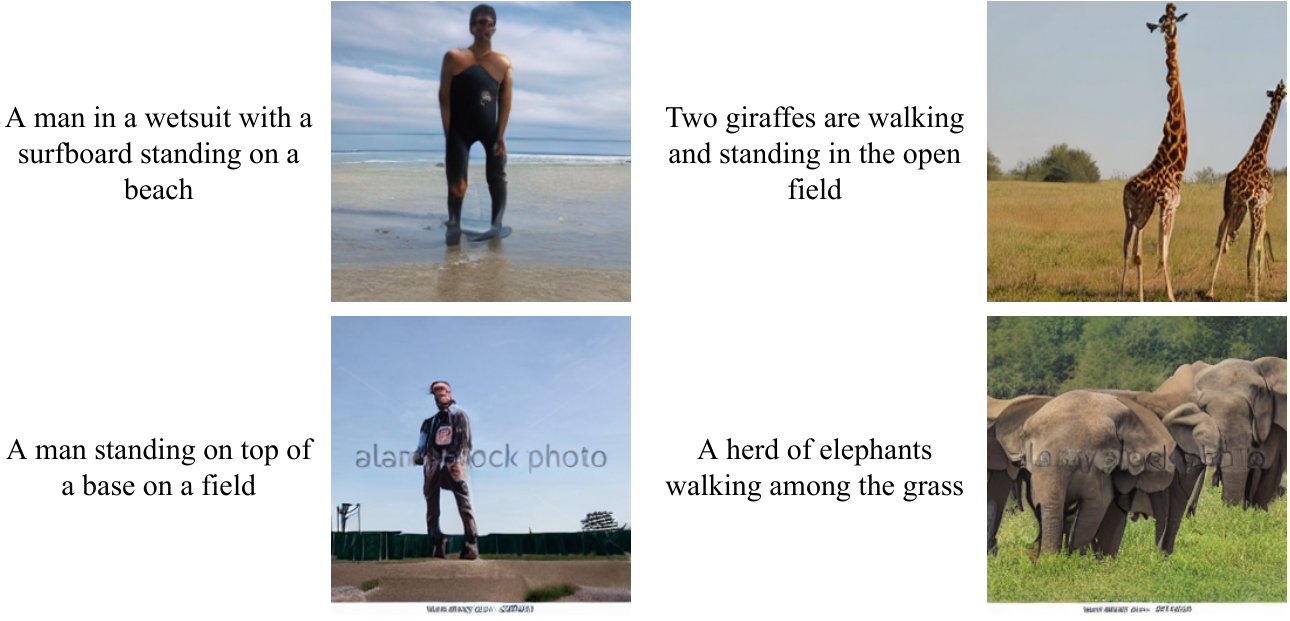}
\end{center}
  \caption{Failure cases of text-to-image generation.}
\label{fig:failure_cases}
\vspace{-2mm}
\end{figure}

The performance of image generation is relatively unsatisfactory (shown in Fig. \ref{fig:failure_cases}) considering the following reasons. 1). Since we train the model on CC3M \cite{sharma2018conceptual}, which contains images of general scenes, the generation quality of some specific domains like humans, animals is low. Training data from these domains may further improve the generation quality (upper). 
2). The generation results may contain watermarks since some images in CC3M are watermarked~(bottom). 

\section{More Implement Details}
In this section, we introduce more implementation details for our DiffDis. 1) We use a cosine noise scheduler for the text query diffusion process and a linear noise scheduler for the image diffusion process. 
2). We assign the timestep of 1000 to the image condition when performing discriminative tasks. Note that 1000 is not in the range of the timestep for image generation. 

Here we give detailed experimental settings for the CLIP models we compared in the main paper. We set the batch size to 1024 and pre-training was conducted for 20 epochs by using AdamW optimizer. The learning rate is 1e-3 and the weight decay is 0.1. During pre-training, the images are randomly cropped and we use the RandAugment \cite{cubuk2020randaugment} for image augmentation. We compare our implementation with open source clip pretraining codebase \cite{ilharco_gabriel_2021_5143773}. We keep the same batch size and the number of training epochs. The experimental results are shown in Table \ref{tab:comparison-open-source-repo-table}. Our implementation is better than open source codebase. We think that the improvement can be attributed to more extensive augmentation for images.

\begin{table}
\center
\setlength{\tabcolsep}{5pt}
\renewcommand\arraystretch{0.92}
\begin{tabular}{lccc}
\toprule
 Codebase & Model & ZS-Acc  \\
\midrule
 OpenCLIP \cite{ilharco_gabriel_2021_5143773} & CLIP-ViT-B/32 & 14.7   \\
 OpenCLIP \cite{ilharco_gabriel_2021_5143773} & CLIP-ViT-L/14 & 19.1   \\
 Our        & CLIP-ViT-B/32 & 16.7  \\
 Our        & CLIP-ViT-L/14 & 21.1  \\
\bottomrule         
\end{tabular}
\vspace{2mm}
\caption{Comparison of our implementation and open source implementation \cite{ilharco_gabriel_2021_5143773}.}
\label{tab:comparison-open-source-repo-table}
\end{table}

\section{More Discussion}



\begin{figure*}[h!]
\begin{center}
\includegraphics[width=1.\linewidth]{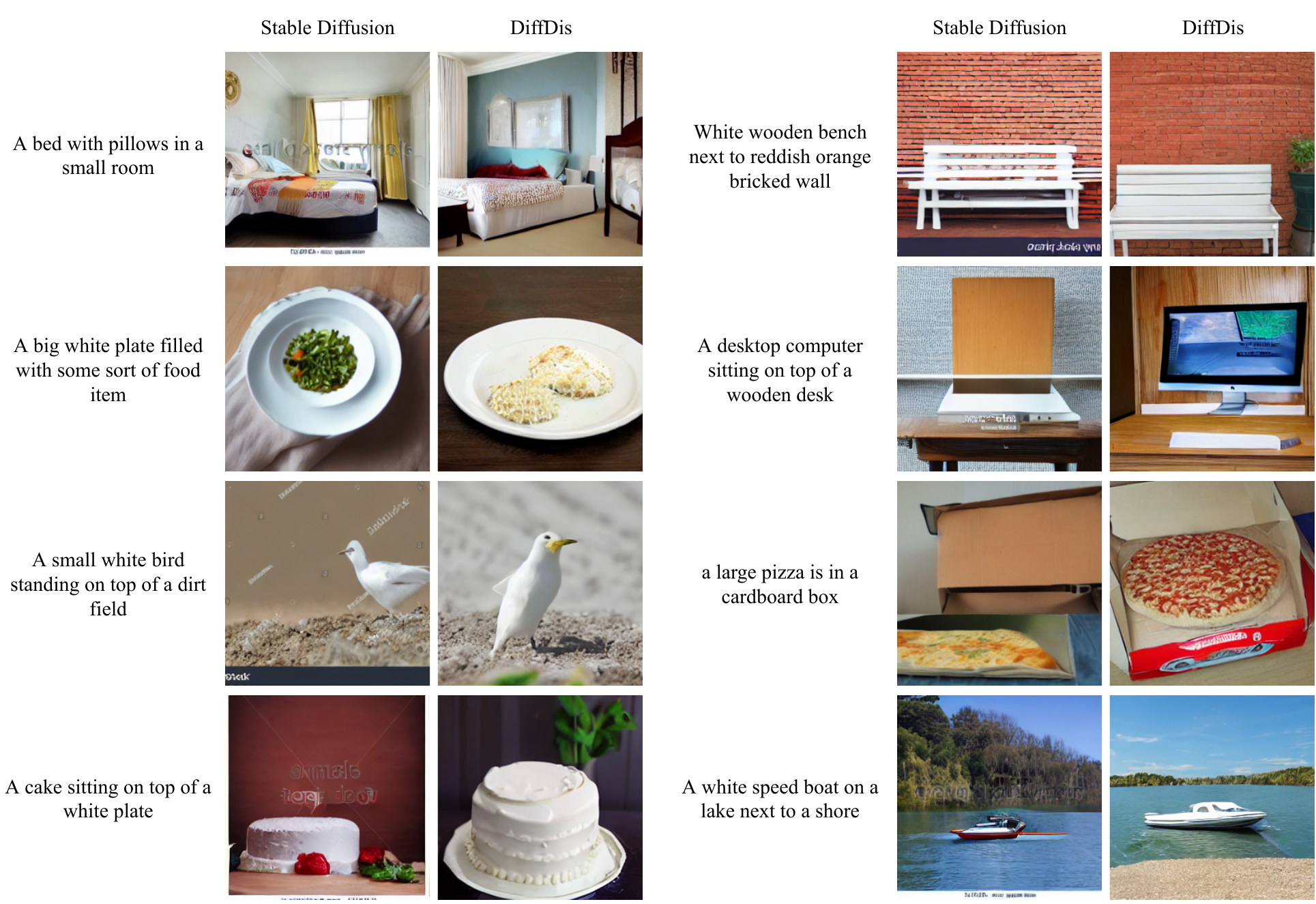}
\end{center}
  \caption{More illustrations of generated samples with proposed DiffDis on MSCOCO prompts.}
\label{fig:more-coco-samples}
\end{figure*}

\noindent\textbf{The Effect of Different Model Targets.} Diffusion model's output can be the original noise $\epsilon$ or the data $x_0$ that denote the noise prediction model and data prediction model, respectively. The comparison of two types of models on three downstream tasks is listed on Table \ref{tab:effect-of-model-target-module}.

\begin{table}
\center
\setlength{\tabcolsep}{5pt}
\renewcommand\arraystretch{0.92}
\begin{tabular}{cccc}
\toprule
Model Target & FID$\downarrow$ & ZS-Acc$\uparrow$ & Mean R@1$\uparrow$  \\
\midrule
Noise & \textbf{10.78}  & 22.62 & \textbf{29.38} \\
Data  & 11.52 & \textbf{23.44} & 28.64 \\
\bottomrule         
\end{tabular}
\vspace{2mm}
\caption{The performance of different model targets. Using feature scaling $\gamma=1$.}
\label{tab:effect-of-model-target-module}
\end{table}

\noindent\textbf{The Effect of Different Noise Schedulers.} We analyze the influence of different noise schedulers on the text diffusion process. The linear schedule starts from 0.00085 to 0.0120. Table \ref{tab:effect-of-noise-scheduler} shows that the linear schedule is a better choice than the cosine schedule. 

\noindent\textbf{The Effect of Dual-Stream Deep Fusion Attention Block.} To evaluate the effectiveness of the proposed dual-stream deep fusion attention block, we disable the fusion block by replacing it with the original attention blocks of Stable Diffusion. We directly concatenate the input text query with the image hidden output from UNet's middle block and feed the concatenation to the 6 blocks transformer. Table \ref{tab:effect-of-fusion-module} shows the experimental results of this comparison. When disabling the deep fusion block, the performances of three downstream tasks are dropped. Besides, according to Table \ref{tab:effect-of-ms-ffn-module}, using modality-specific FFN and sharing the attention module in dual-stream deep fusion attention block will improve the performance on generation tasks. 

\begin{table}
\center
\setlength{\tabcolsep}{5pt}
\renewcommand\arraystretch{0.92}
\begin{tabular}{cccc}
\toprule
Noise Scheduler &  FID$\downarrow$ & ZS-Acc$\uparrow$ & Mean R@1$\uparrow$  \\
\midrule
Cosine & 11.90 & 22.35 & 28.59  \\
Linear & \textbf{11.52} & \textbf{22.70} & \textbf{31.07}  \\
\bottomrule         
\end{tabular}
\vspace{2mm}
\caption{The performance of different noise schedulers. Using feature scaling $\gamma=1$.}
\label{tab:effect-of-noise-scheduler}
\end{table}

\begin{table}
\center
\setlength{\tabcolsep}{5pt}
\renewcommand\arraystretch{0.92}
\begin{tabular}{cccc}
\toprule
Enabled Fusion &  FID$\downarrow$ & ZS-Acc$\uparrow$ & Mean R@1$\uparrow$ \\
\midrule
\xmark & 10.05 & 24.53 & 32.97 \\
\cmark & \textbf{9.80} & \textbf{25.92} & \textbf{33.60} \\
\bottomrule         
\end{tabular}
\vspace{2mm}
\caption{The performance on FID score on MSCOCO image generation, zero-shot ImageNet classification and average R@1 of MSCOCO and Flickr30k by enabling dual-stream deep fusion attention block.}
\label{tab:effect-of-fusion-module}
\end{table}

\begin{table}
\center
\setlength{\tabcolsep}{5pt}
\renewcommand\arraystretch{0.92}
\begin{tabular}{ccccc}
\toprule
 Share Attn & MS-FFN &  FID$\downarrow$ & ZS-Acc$\uparrow$ & Mean R@1$\uparrow$ \\
\midrule
\cmark & \xmark &  10.26 & \underline{25.92} & \textbf{33.75} \\
\xmark & \cmark & 10.19 & \textbf{26.25} & 33.07 \\
\cmark & \cmark & \textbf{9.80} & \underline{25.92} & \underline{33.60}\\
\bottomrule         
\end{tabular}
\vspace{1mm}
\caption{The effect of the modality-specific FFN (MS-FFN) and sharing attention module in the dual-stream deep fusion attention block. We use the setting of the last row in our model.}
\label{tab:effect-of-ms-ffn-module}
\end{table}

\noindent\textbf{Time Comparison.} We provide the training time, generative inference time on COCO and discriminative inference time on ImageNet in Table \ref{tab:comparison-training-inference-time-table}. After unifying the discriminative and generative tasks, DiffDis has a longer training time compared to single-task training but has a shorter training time than
the sum training time of CLIP-ViT-L/14 and Stable Diffu-
sion and make better or comparable performance. DiffDis
has a similar generative inference time as Stable Diffusion
and 1.7x discriminative inference time compared to CLIP.

\begin{table}
\center
\setlength{\tabcolsep}{5pt}
\renewcommand\arraystretch{0.92}
\Large
\resizebox{\linewidth}{!}{
\begin{tabular}{lccccc}
\toprule
 Time / Tasks       & Training   & Gen-Inference & Dis-Inference & ZS-Acc$\uparrow$  & FID$\downarrow$ \\
\midrule
 CLIP-ViT-L/14    & 1d 7h      &  --      &  148s   & 21.1  & --   \\
 Stable Diffusion & 1d 8h      & 3530s    &  --     &  --   & 10.8 \\
 DiffDis          & 2d 6h      & 3550s   &  252s   & \textbf{25.9}  & \textbf{9.8}  \\
\bottomrule         
\end{tabular}
}
\vspace{2mm}
\caption{The training time and inference time comparison. }
\label{tab:comparison-training-inference-time-table}
\end{table}



\noindent\textbf{The Mask Timestep of Image Condition for the Discriminative Tasks.}~~The image condition for the discriminative tasks needs a timestep to input. We discuss the selection of the image condition on three downstream tasks on Table \ref{tab:comparison-mask-timestep-for-image-condition-table}. The experimental results show that reusing the timestep within the range of image generation's timestep leads to performance degradation on both image generation tasks and discriminative tasks. The use of the `First' mask timestep ($t_z=0$) will degrade the performance most. Assigning an additional timestep for the  image condition for discriminative tasks achieves the best performance on all downstream tasks.

\begin{table}
\center
\setlength{\tabcolsep}{5pt}
\renewcommand\arraystretch{0.92}
\begin{tabular}{lcccc}
\toprule
Position & $t_z$ & FID$\downarrow$ & ZS-Acc$\uparrow$ & Mean R@1$\uparrow$ \\
\midrule
First & 0          & 12.35  & 21.97          & 27.20 \\
Last  & 999        & 12.02  & 21.73          & \textbf{27.56}  \\
Additional & 1000  & \textbf{11.35}  & \textbf{22.13} & \textbf{27.56} \\
\bottomrule         
\end{tabular}
\vspace{2mm}
\caption{Results of different mask timestep of image condition for discriminative learning. The range of the image generation diffusion steps is 0-999 . The additional timestep used for discriminative tasks is not shared with image generation.}
\label{tab:comparison-mask-timestep-for-image-condition-table}
\end{table}

\noindent\textbf{Discussion with HybViT} We clarify that the DiffDis cannot directly compare with HybViT~\cite{Yang2022HibViT} since 1) HybViT focuses on class-condition image generation while our DiffDis targets text-condition image generation; 2) HybViT performs supervised classification tasks but can not perform zero-shot classification tasks or image-text retrieval tasks while DiffDis can.

\section{Application of DiffDis}
We follow VL-LTR~\cite{tian2021vl} to perform long-tailed visual recognition tasks and apply DiffDis or CLIP-ViT-L/14~(our implementation, pre-trained on CC3M), as the backbone. As shown in Table \ref{tab:comparison-of-long-tailed-recognition-table} We evaluate the performances of the pre-train stage and fine-tune stage on the Places-LT dataset~\cite{liu2019large}. 
\begin{table}
\center
\setlength{\tabcolsep}{5pt}
\renewcommand\arraystretch{0.92}
\resizebox{\linewidth}{!}{
\begin{tabular}{l|ccc|c}
\toprule
\multirow{2}{*}{Backbone}  &  \multicolumn{3}{c|}{Pre-train Stage}    &  Fine-tune Stage \\
                          &     Image-Acc   & Text-Acc & KNN-Acc      & Acc \\
\midrule
 CLIP-ViT-L/14         &  31.4          & 38.1         &35.5          & 40.5   \\
 DiffDis               &  \textbf{37.0} & \textbf{52.5}&\textbf{40.5} &  \textbf{44.4}   \\
\bottomrule         
\end{tabular}
}
\vspace{2mm}
\caption{Results of long-tailed recognition on Places-LT dataset by using different backbone. We follow the official code of VL-LTR~\cite{tian2021vl}.}
\label{tab:comparison-of-long-tailed-recognition-table}
\end{table}

\end{document}